\documentclass[nohyperref]{article}

\usepackage{microtype}
\usepackage{graphicx}
\usepackage{subfigure}
\usepackage{booktabs} 

\usepackage{hyperref}

\newcommand{\pan}[1]{\textcolor{blue}{(Pan: #1)}}

\usepackage{amsmath}
\usepackage{amssymb}
\usepackage{mathtools}
\usepackage{amsthm}
\usepackage{enumitem}
\usepackage{comment}

\usepackage{multirow}
\usepackage{xspace}
\newcommand{\proj}{StruRW\xspace}
\newcommand{\projad}{StruRW-Adv\xspace}
\newcommand{\projmix}{StruRW-Mix\xspace}
\newcommand{\projerm}{StruRW-ERM\xspace}

\usepackage[accepted]{icml2023}

\usepackage{booktabs}
\usepackage{wrapfig,lipsum}
\usepackage{graphicx}
\usepackage{algorithmic}
\usepackage{caption}
\usepackage{wrapfig}

\usepackage{xspace}
\usepackage{booktabs}
\usepackage{xcolor}         
\usepackage{comment}
\usepackage{enumitem}
\usepackage{wrapfig}
\usepackage{setspace}
\usepackage{amsthm}
\usepackage{amsmath}
\usepackage{array,multirow,graphicx}
\usepackage{amssymb}
\usepackage{mathtools}
\usepackage{adjustbox}

\usepackage[capitalize,noabbrev]
{cleveref}

\theoremstyle{plain}
\newtheorem{theorem}{Theorem}[section]
\newtheorem{proposition}[theorem]{Proposition}

\theoremstyle{definition}
\newtheorem{definition}[theorem]{Definition}

\newtheorem{example}[theorem]{Example}
\theoremstyle{remark}

\newcommand*{\ldbb}{\{\mskip-5mu\{}
\newcommand*{\rdbb}{\}\mskip-5mu\}}
\usepackage[textsize=tiny]{todonotes}

\newcommand{\dG}{\mathcal{G}}
\newcommand{\dS}{\mathcal{S}}

\newcommand{\dV}{\mathcal{V}}
\newcommand{\dE}{\mathcal{E}}

\newcommand{\dX}{\mathcal{X}}
\newcommand{\dT}{\mathcal{T}}
\newcommand{\dU}{\mathcal{U}}
\newcommand{\dN}{\mathcal{N}}
\newcommand{\dH}{\mathcal{H}}
\newcommand{\dY}{\mathcal{Y}}
\newcommand{\mX}{\mathbf{X}}
\newcommand{\mY}{\mathbf{Y}}
\newcommand{\mA}{\mathbf{A}}
\newcommand{\mB}{\mathbf{B}}
\newcommand{\mH}{\mathbf{H}}

\newcommand{\mx}{\mathbf{x}}
\newcommand{\my}{\mathbf{y}}
\newcommand{\mh}{\mathbf{h}}

\newcommand{\mPi}{\boldsymbol{\Pi}}

\newcommand{\bP}{\mathbb{P}}

\icmltitlerunning{Structural Re-weighting Improves Graph Domain Adaptation}

\begin{document}

\twocolumn[
\icmltitle{Structural Re-weighting Improves Graph Domain Adaptation}



\icmlsetsymbol{equal}{*}

\begin{icmlauthorlist}
\icmlauthor{Shikun Liu}{gt}
\icmlauthor{Tianchun Li}{purdue}
\icmlauthor{Yongbin Feng}{fermi}
\icmlauthor{Nhan Tran}{fermi}
\icmlauthor{Han Zhao}{uiuc}
\icmlauthor{Qiu Qiang}{purdue}
\icmlauthor{Pan Li}{gt}
\end{icmlauthorlist}

\icmlaffiliation{gt}{Department of Electrical and Computer Engineering, Georgia Institute of Technology, Georgia, U.S.A}
\icmlaffiliation{purdue}{Department of Electrical and Computer Engineering, Purdue University, West Lafayette, U.S.A}
\icmlaffiliation{fermi}{Fermi National Accelerator Laboratory, Batavia, U.S.A}
\icmlaffiliation{uiuc}{Department of Computer Science, University of Illinois Urbana-Champaign, Champaign, U.S.A}

\icmlcorrespondingauthor{Shikun Liu }{shikun.liu@gatech.edu}
\icmlcorrespondingauthor{Pan Li}{panli@gatech.edu}

\icmlkeywords{Machine Learning, ICML}

\vskip 0.3in
]



\printAffiliationsAndNotice{} 

\begin{abstract}
In many real-world applications, graph-structured data used for training and testing have differences in distribution, such as in high energy physics (HEP) where simulation data used for training may not match real experiments. Graph domain adaptation (GDA) is a method used to address these differences. However, current GDA primarily works by aligning the distributions of node representations output by a single graph neural network encoder shared across the training and testing domains, which may often yield sub-optimal solutions. This work examines different impacts of distribution shifts caused by either graph structure or node attributes and identifies a new type of shift, named conditional structure shift (CSS), which current GDA approaches are provably sub-optimal to deal with. A novel approach, called  structural reweighting (\proj), is proposed to address this issue and is tested on synthetic graphs, four benchmark datasets, and a new application in HEP. \proj has shown significant performance improvement over the baselines in the settings with large graph structure shifts, and reasonable performance improvement when node attribute shift dominates. \footnote{Our code is available at: \url{https://github.com/Graph-COM/StruRW}}
\end{abstract}

\vspace{-2mm}
\section{Introduction}
\label{sec:intro}
Graph neural networks (GNNs) have recently become the de facto tool to learn the representations of graph-structured data~\cite{scarselli2008graph,kipf2017semisupervised}. Despite their exceptional performance on benchmarks~\cite{hu2020open,hu2021ogb}, GNNs have been found to struggle in high-stakes real-world applications where there is a data-distribution shift between the training and test phases~\cite{li2022critical,hu2020strategies,gaudelet2021utilizing}. 

\begin{figure}[t]
\begin{center}
\centerline{\includegraphics[trim={0.83cm 0.15cm 0.15cm 0.18cm},clip,width=0.93\columnwidth]{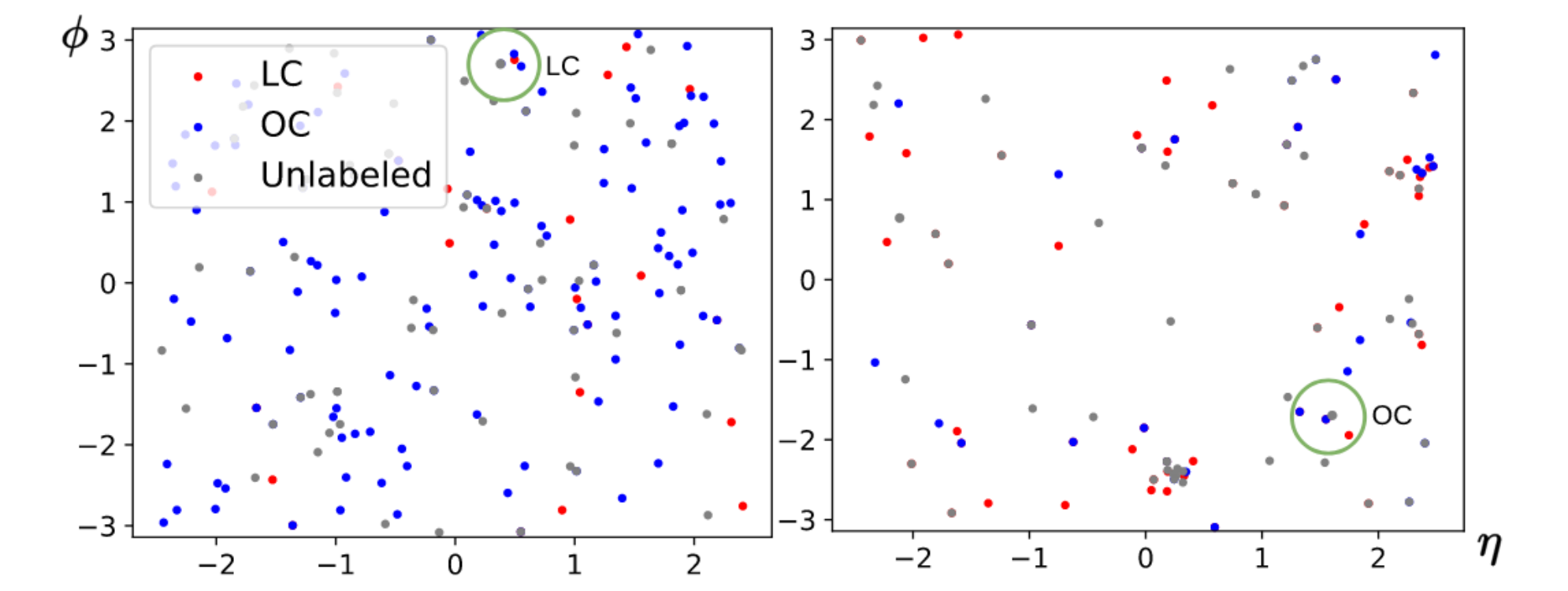}}
\caption{\small{Examples of pileup events in two different PU levels: PU30 (Left) and PU10 (Right). Charged particles can be labeled as LC or OC, as highlighted by red or blue dots, while the labels of neutral particles are often unknown. Graph methods are often used here~\cite{bertolini2014pileup,li2022semi}, where KNN graphs can be built and one can leverage nearby particle features and labels to make inference, highlighted by e.g. the two circles in the figure. In either of the two circles, there are two 2 labeled OC's and 1 labeled LC, while the ground-truth label of the neutral particle in the center may be different from each other.} }
\label{fig:HEPeg}
\end{center}
\vspace{-8.5mm}
\end{figure}

This study is motivated by applications in high energy physics (HEP)~\cite{shlomi2020graph}, where GNNs are often trained on simulated data with an abundance of labels and then applied to real experiments with limited labels~\cite{nachman2019ai}. However, real experiments have complex, time-varying environments that may differ from simulated setups. One such example is the change in pile-up (PU) levels in Large Hadron Collider (LHC) experiments~\cite{highfield2008large}. PU level refers to the number of collisions around the main collision of interest, which can change over time and differ from the levels used to generate simulation data. Modeling the data using graphs, the connection patterns between particles in different PU levels will significantly change, as depicted in Fig.~\ref{fig:HEPeg}. This poses a major challenge for GNNs to distinguish particles from the leading collision (class LC) from those from other collisions (class OC), which is a crucial task in HEP data analysis~\cite{perloff2012pileup,li2022semi}. Similar shifts also occur in social and biological networks, where the interaction patterns between nodes with different labels can change over time~\cite{wang2021inductive} or across different species~\cite{cho2016compact}, as listed in Table~\ref{table:CSS}.  

Graph domain adaptation (GDA) has been proposed to deal with such distribution shift problems. Current GDA methods frequently utilize GNNs as a means of creating dense node representations, and then implement regularization in order to ensure these representations remain consistent across both the training (source) and test (target) domains~\cite{wu2020unsupervised, xiao2022domain,zhu2021shift}. However, this approach largely overlooks the distinct effects of distribution shifts caused by graph structures and node representations, and as a result, may not yield optimal solutions.

\begin{table}[t]
\caption{\small{Conditional Structure Shift (CSS, computed according to Eq.~\eqref{eq:CSS} ) across real datasets that are used for evaluation in this work. CSS will be explained in Sec.~\ref{subsec:struc-cond-shift}.}}
\resizebox{\columnwidth}{!}{%
\begin{tabular}{lccccccc}
\toprule
 & \multicolumn{2}{c}{DBLP and ACM} & \multicolumn{2}{c}{Cora} &   \multicolumn{3}{c}{Arxiv}\\
 \toprule
Domains  &   $A\rightarrow D$    & $D \rightarrow A$ & Word &  Degree & Time1      &     Time2 &  Degree      \\
\midrule
$\widehat{CSS}$ & $7.4276$ & $7.4276$ & $0.5583$ & $0.9980$& $1.0106$ & $1.2148$ & $2.6131$\\
\bottomrule
\label{table:CSS}
\end{tabular}%
}
\vspace{-5mm}
\end{table}

In this work, we investigate different types of distribution shifts of graph-structured data and offer significant understanding into GDA for \underline{node classification} problems. 
First, we show that if the objective is to acquire node representations with distributions that remain invariant across domains, adding regularization to the last-layer node representations is adequate. Imposing regularization on intermediate node representations or matching node initial attributes across two domains may actually induce extra loss. 


Though with the above observation, we further show that it is suboptimal in many cases to achieve such distribution invariance via a single stand-alone GNN encoder shared across domains. To illustrate the problem, we revisit the HEP example in Fig.~\ref{fig:HEPeg}: when the PU level is high (PU30), an unlabeled particle that is connected to one LC particle and two OC particles is more likely to be classified as LC. Conversely, in instances where the PU level is low (PU10), the particle with the same neighborhood may be more likely to be classified as LC due to the expectation of more OC particles in the vicinity of an OC particle. Under these scenarios, the optimal node representations with the same neighborhood should actually change to fit different domains rather than keep invariant. In this work, we formally define this new type of distribution shift as conditional structure shift (CSS). The CSS not only exists under the HEP setting but in other real applications, like social networks. For instance, different periods of time in citation networks may present different citation relations across fields due to the change of focus on interdisciplinary work or related work over time. We will discuss the detailed degree of CSS with other real datasets in Section ~\ref{subsec:struc-cond-shift}. Current GDA methods fail to address CSS properly. 

\begin{figure}[t]
\begin{center}
\centerline{\includegraphics[trim={1.2cm 9.2cm 14.5cm 1.8cm},clip,width=1\columnwidth]{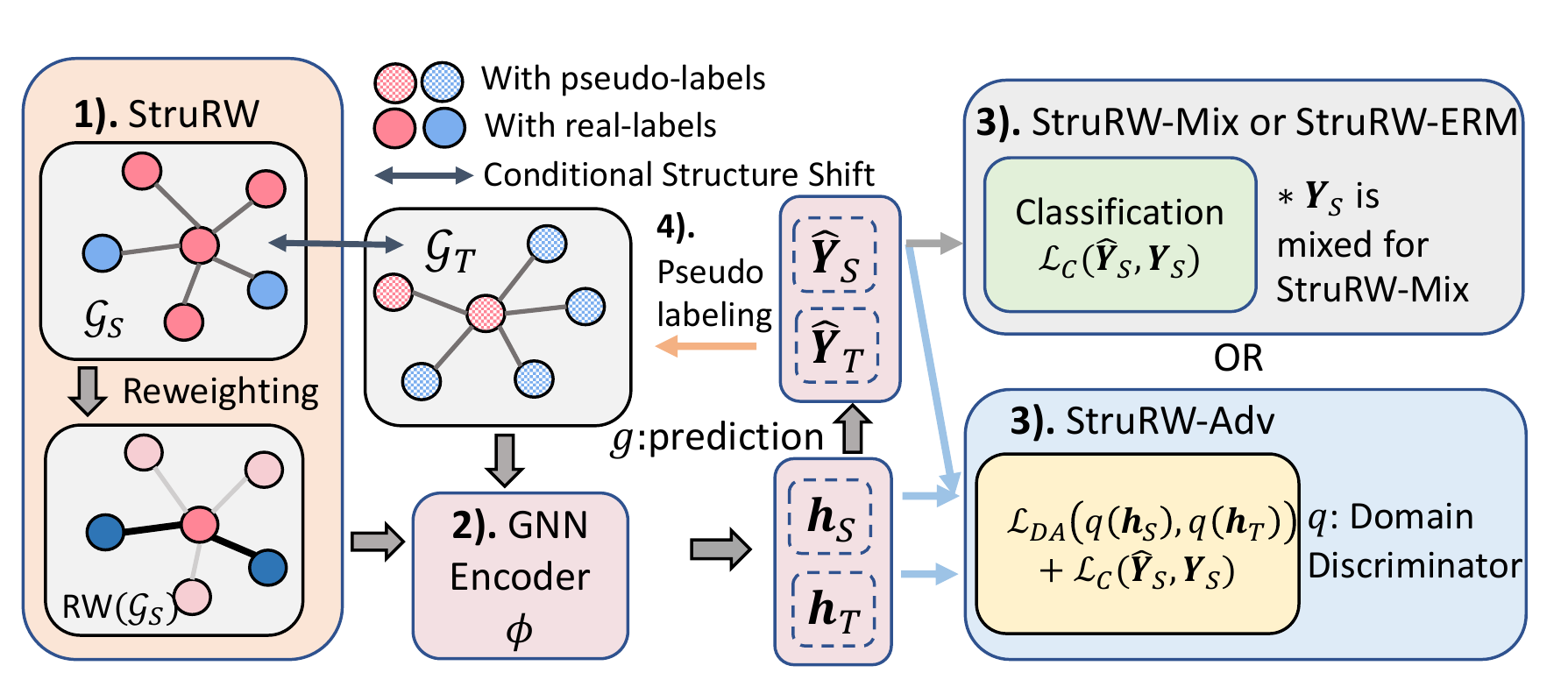}}
\vspace{-3mm}
\caption{\small{This diagram demonstrates the model pipeline by combining the \proj module, GNN encoder, then the generalized loss calculation block that supports \projad, \projmix, and \projerm. The pseudo codes are presented in Algorithm~\ref{method:alg_table}.}}
\label{fig:pipeline}
\end{center}
\vspace{-10mm}
\end{figure}

To deal with CSS, we propose a novel GDA algorithm named structural re-weighting (\proj) as shown in Fig. ~\ref{fig:pipeline}. \proj computes the edge probabilities between different classes based on the pseudo node labels estimated on the target graphs, and then uses these probabilities to guide bootstrapping of neighbors used in GNN computation on the source graphs, which eventually reduces the conditional shift of neighborhoods. The GNN composed with \proj differentiates the encoding processes across the domains, which breaks the limitation.

We conduct extensive experiments on synthetic graphs, four real-world benchmarks and one HEP dataset to verify our theory and the effectiveness of \proj. 
Across the cases, \proj has achieved significant improvements over baselines under the settings with obvious graph structure shifts, and slight improvements for other settings dominated by node attribute shifts. Due to the page limitation, we leave the proofs of all propositions in this work in the appendix.
\vspace{-1mm}
\section{Preliminaries and Related Works}  
\label{sec:prelim}

In this section, we introduce basic concepts and notations to set up the problem and review related works along the way. 

\textbf{Domain Adaptation (DA).}
This work studies unsupervised DA, where the model has access to labeled data 
from the source domain and unlabeled data 
from the target domain, and the goal is to train the model to achieve small classification error on the target domain. 

To review the general idea of DA methods, denote $\bP_X$ as the distribution of the feature $x\in\dX$. We always use subscripts/superscripts $\dU\in\{\dS,\dT\}$ to denote the source and target domains respectively. 
Denote $f_\dU$ as the true labeling function that maps $x$ to labels $y\in\dY$ for domain $\dU$. For simplicity, we temporarily assume binary classification $\dY=\{0,1\}$ to show theoretical insights, while the proposed algorithm can be applied to other cases. Suppose the model has a composition form $g\circ \phi$ that first maps the features to a latent space $\phi:\dX\rightarrow \dH$ and then performs classification $g:\dH\rightarrow \dY$. Then, the classification error of the model in domain $\dU$ can be denoted as $\epsilon_\dU(g,\phi) = \mathbb{E}_{x\in \mathbb{P}_X^\dU}[|g(\phi(x))-f_\dU(x)|]$. By adopting a derivation similar to \cite{wu2019domain, ben2010theory}, the error in the target domain can be bounded as follows. The detailed derivation is shown in Appendix~\ref{adp:bound}. \\
\vspace{-2mm}
\begin{align}
    \epsilon_{\dT}&(g,\phi) \leq \epsilon_{\dS}(g,\phi) + \int_{h} d\bP^\dT_{X}(h)\left| f_{\dS}^\phi(h) - f_{\dT}^\phi(h)\right| \nonumber \\
    &\quad  +  \int_h |d\mathbb{P}_\phi^\dT(h) - d\mathbb{P}_\phi^\dS(h)r_\dS(h, \phi, g) \label{eq:bound}
\end{align}
where $r_\dU(h, \phi, g) \triangleq \int_{x:\phi(x) = h}|g(h) - f_{\dU}(x)|d\bP_X^\dU(x)$, $f_{\dU}^\phi(h) \triangleq \int_{x:\phi(x) = h}f_{\dU}(x)d\bP_X^\dU(x)$ is the labeling function from the latent space, and $d\bP^\dU_{\phi}(h) = \int_{x:\phi(x)=h} d\bP_X^\dU(x)$. 

To minimize the target error, one common way in DA is to \emph{push the encoder $\phi$ to output representations with the distribution invariant across domains} by minimizing the third term while minimizing the source error, i.e., the first term. 
The second term is often overlooked as it is hard to control. 


Previous methods to learn invariant representations adopt some regularization methods, including adversarial training with domain discriminator~\cite{ganin2016domain, zellinger2017central}, or minimizing some distribution-distance measures~\cite{long2015learning} such as Maximum Mean Discrepancy (MMD)~\cite{ saito2018maximum}  between the source and target latent representations.

\textbf{Graph Neural Networks (GNNs).} 
Let $\dG=(\dV,\dE,\mx)$ denote an undirected graph with a node set $\dV$, an edge set $\dE$ and node attributes $\mx=[\cdots x_v\cdots]_{v\in \dV}$. The graph structure can also be denoted as the adjacency matrix $\mA$ where its entry $A_{uv}=1$ if edge $uv\in\dE$ and otherwise $A_{uv}=0$. 
GNNs encode $\mA$ and $\mx$ into node representations $\{h_v|v\in \dV\}$. Initialize $h_v^{(0)}=x_v$ and standard GNNs~\cite{hamilton2017inductive,gilmer2017neural} follow a message passing procedure. Specifically, for each node $v$ and for $l=0,1,...,L-1$,
\begin{equation}
    h_v^{(l+1)}=\text{UDT}\,(h_v^{(l)}, \text{AGG}\,(\ldbb h_u^{(l)}:u \in \dN_v\rdbb)), 
    \label{eq:GNN}
\end{equation}
where $\dN_v$ denotes the set of neighbors of node $v$ and $\ldbb\cdot\rdbb$ denotes a multiset. The AGG function aggregates messages from the neighbors, and the UPT function updates the node representations. In \emph{node classification} tasks, the last-layer node representation $h_v^{(L)}$ is used to predict the label $y_v\in\dY$. 



\textbf{Graph Domain Adaptation (GDA).} GDA extends DA to the setting with graph-structured data.   Specifically, we have one or several graphs $\dG_{\dS} = (\dV_{\dS}, \dE_{\dS}, \mx_{\dS})$ from the source domain with node labels $\my_{\dS}$ and one or several graphs $\dG_{\dT} = (\dV_{\dT}, \dE_{\dT}, \mx_{\dT})$ from the target domain. The goal is to predict node labels $\my_{\dT}$ in the target domain. Different from traditional DA with independent data points,  features and labels are coupled due to the graph structure. Existing graph methods address the problem by first adopting a GNN to encode the graph into node representations $\mh^{(L)}=[\cdots h_v^{(L)}\cdots]_{v\in\dV}$, and then enforcing invariance on the representations in $\mh^{(L)}$ across domains. 

\textbf{Related Works.} For the related works with specific implementations of above GDA idea, DANE~\cite{zhang2019dane} introduces adversarial training of domain classifier based on those node representations. UDAGCN~\cite{wu2020unsupervised} further imposes some inter-graph attention mechanism on top of the adversarial training. SR-GNN~\cite{zhu2021shift} aims to minimize the moment distance between the node-representation distributions across domains. DGDA~\cite{cai2021graph} aims to disentangle semantic, domain, and noise variables and uses semantic variables that are better aligned with target graphs for prediction. 
All these works did not analyze the potential distribution shifts for node classification tasks and may therefore suffer from the CSS problem.
A very recent work~\cite{you2023graph} proposes to use graph spectral regularization to address GDA problems. Although this work extends the generalization bound in ~\cite{zhao2019learning} for the case with the conditional shift in the scenario of GDA, their algorithm is not designed to address the issue of conditional shift. 



In addition to GDA, many works aim to train GNNs for out-of-distribution  (OOD) generalization. Different from GDA, they do not assume the availability of unlabeled test data and expect to train a GNN that learns representations invariant to generic domain change. Hence, they cannot address the problem in Fig.~\ref{fig:HEPeg} as well. For node classification tasks, EERM~\cite{wu2022handling} minimizes the variance of representations across different generated environments. \citet{ma2019disentangled} and \citet{liu2020independence} extract invariant features by disentangling the entries of node representations. 
\citet{verma2021graphmix,wang2021mixup} mixup node representations across different classes for training to flatten the decision boundary~\cite{zhang2018mixup}. 
\citet{qiu2020gcc,wu2022knowledge,park2021metropolis,liu2022local,you2020graph} adopt data augmentation to achieve betteer generalization. 
Other works study OOD graph classification tasks and can be categorized similarly as above~\cite{zhu2021transfer, miao2022interpretable, chenlearning, li2022learning, han2022g, yang2022learning,suresh2021adversarial}.   

\textbf{Other Notations} In the following, we use capital letters e.g., $\mX,X$ to denote random variables (r.v.) and the lower-case letters, e.g., $\mx, x$ to denote specific values, except the adjacency matrix $\mA$ that will be used to denote both. Use $\mPi$ to denote a permutation matrix with a proper dimension. 
\section{Optimality of Last-layer Domain Invariance}\label{sec:analysis}




In this section, we disentangle the types of distribution shifts in graph-structured data and look into the question of whether regularizing only the last-layer node representations, as commonly adopted, is optimal to learn node representations invariant across domains under various types of shifts.


\subsection{Distribution Shifts in Graph-structured Data} \label{subsec:shift-def}
We categorize different types of distribution shifts in graph-structured data for node classification problems.

\textbf{Structure shift.} 
Consider the joint distribution of the adjacency matrix and node labels $\bP_{\mA\times \mY}$. Structure distribution has internal symmetry where $\bP_{\mA\times \mY}(\mA, \my) = \bP_{\mA\times \mY}(\mPi\mA\mPi^\top, \my)$ for any $\mPi$ s.t. $\my=\mPi\my$. \underline{Structure shift} is defined for the case when $\bP_{\mA\times \mY}^\dS\neq\bP_{\mA\times \mY}^\dT$. 

\textbf{Attribute shift.} We assume that without the graph structure, the attributes $x_v$, $v\in\dV$ are IID sampled from $\bP_{X|Y}$ given node labels $y_v$. Therefore, the conditional distribution of $\mx|\my$ satisfies $\bP_{\mX|\mY}(\mx|\my) =\prod_{v\in \dV} \bP_{X|Y}(x_v|y_v)$, which satisfies $\bP_{\mX|\mY}(\mx|\my)= \bP_{\mX|\mY}(\mPi\mx|\my)$ for any $\mPi$ such that $\mPi\my=\my$. Then, \underline{Attribute shift} refers to $\bP^\dS_{X|Y}\neq \bP^\dT_{X|Y}$.

We use the joint distribution to define structure shift while the conditional distribution to define attribute shift because it better aligns with practice: Graph structure captures the correlation between nodes including their labels while node attributes are often independent given their labels. 

\subsection{Analysis for GDA with Different Types of Shifts} \label{subsec:observation}

Our analysis is built upon the error bound in Eq.~\eqref{eq:bound} that reveals the goal of learning domain-invariant node representations while minimizing the error in the source domain $\epsilon_{\dS}(g,\phi)$. 
For GDA, the GNN is denoted as $\phi$ to transform the graph into node representations $\mh^{(L)}=\phi(\mx,\mA)$ and the downstream node classifier is $g$. 
Note that in GDA, the entries of $\mh^{(L)}$ are not independent of each other. 
The common practice to deal with this issue is to use a sampling procedure to marginalize the joint distribution:

\begin{definition}[Marginalization] For domain $\dU$, given node representations $\mh^{(l)}$, marginalization is to uniformly sample one of them $h_v^{(l)}$. Denote the distribution of $h_v^{(k)}$ as $\mathbb{P}_{\phi}^{\dU}$.
\end{definition}

With marginalization, the goal of learning domain-invariant node representations for GDA can be reduced to  
\begin{equation}
    \label{eq:analysis-goal}
   \min_{g,\phi}\; \epsilon_{\dS}(g,\phi) \quad \text{s.t.} \; \mathbb{P}_{\phi}^{\dS} = \mathbb{P}_{\phi}^{\dT}.
\end{equation}

We break the GNN into two parts $\phi=\phi_{>l}\circ\phi_{\leq l}$ where $\phi_{\leq l}$ denotes the encoder of the first $l(<L)$ layers $\mh^{(l)}=\phi_{\leq l}(\mx,\mA)$ and $\mh^{(L)}=\phi_{>l}(\mh^{(l)},\mA)$. 
With some abuse of notation, let $\phi_{\leq0}$ denote the first-layer transformation of node attributes before passing them to the neighbors. We use $\mathbb{P}_{\phi_{\leq l}}^{\dS} = \mathbb{P}_{\phi_{\leq l}}^{\dT}$ to indicate that the distributions of the marginalization of $\mh^{(l)}$ are invariant across domains.  

Given these notations, our question reduces to whether imposing $\mathbb{P}_{\phi}^{\dS} = \mathbb{P}_{\phi}^{\dT}$ is optimal for Eq.~\eqref{eq:analysis-goal} and whether imposing $\mathbb{P}_{\phi_{\leq l}}^{\dS} = \mathbb{P}_{\phi_{\leq l}}^{\dT}$ for some $l<L-1$ can be better. We consider two cases with or without structure shift by assuming there always exists of attribute shift because otherwise structure shift can be transformed into a shift of node representations (similar to attribute shift).

\textbf{Case I: Without structure shift.} As we only have attribute shift in this case, an interesting question is whether aligning the distributions of node attributes can do better since the structure has no shift. 

\begin{wrapfigure}{r}{0.24\textwidth}
    \centering
    \vspace{-8mm}
    \includegraphics[scale=0.22,width=0.24\textwidth]{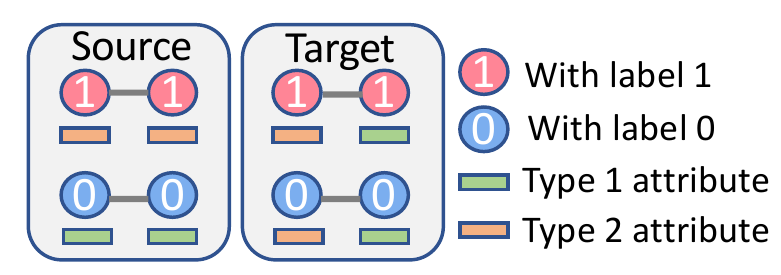}
    \vspace{-7mm}
    \caption{An example for $\mathbb{P}_{\phi_{\leq 0}}^{\dS} = \mathbb{P}_{\phi_{\leq 0}}^{\dT} \not \Rightarrow\mathbb{P}_{\phi}^{\dS} = \mathbb{P}_{\phi}^{\dT}$. }
    \label{fig:sec4eg}
    \vspace{-3mm}
\end{wrapfigure}

First, we argue that just aligning the distributions of node attributes $\mathbb{P}_{\phi_{\leq 0}}^{\dS} = \mathbb{P}_{\phi_{\leq 0}}^{\dT}$ is insufficient to achieve final invariance $\mathbb{P}_{\phi}^{\dS} = \mathbb{P}_{\phi}^{\dT}$ even without structure shift. This can be illustrated with an example shown in Fig.~\ref{fig:sec4eg}: The marginal distribution of node attributes are the same across the domains $\mathbb{P}_{\phi_{\leq 0}}^{\dS} = \mathbb{P}_{\phi_{\leq 0}}^{\dT}$ and there is no structure shift. However, after one layer of GNN, there will be a distribution shift in node representations. 

Second, as shown in Proposition~\ref{prop:remove-attribute-shift}, aligning the conditional distributions of node attributes $\mathbb{P}_{\phi_{\leq 0}|Y}^{\dS} = \mathbb{P}_{\phi_{\leq 0}|Y}^{\dT}$ may be sufficient under some independence assumption. This seems to give a chance to outperform previous methods that impose $\mathbb{P}_{\phi}^{\dS} = \mathbb{P}_{\phi}^{\dT}$ in the last layer. 



\begin{proposition}
\label{prop:remove-attribute-shift}
    Suppose the node attributes and the graph structures are independent given the node labels in the two domains $\bP_{(\mX,\mA)|\mY}^{\dU}(\mX,\mA|\my)=\bP_{\mX|\mY}^{\dU}(\mX|\my)\bP_{\mA|\mY}^{\dU}(\mA|\my)$. If there is no structure shift $\bP_{\mA,\mY}^{\dS}(\mA,\my)=\bP_{\mA,\mY}^{\dT}(\mA,\my)$, a transformation $\phi_{\leq 0}$ of the node attributes that can achieve $\mathbb{P}_{\phi_{\leq 0}|Y}^\mathcal{S} = \mathbb{P}_{\phi_{\leq 0}|Y}^\mathcal{T}$ is sufficient to make the distributions of last-layer node representations invariant across domains, i.e.,  $\mathbb{P}_\phi^\mathcal{S} = \mathbb{P}_\phi^\mathcal{T}$ without the need of further regularization. 
\end{proposition}

 However, we hardly see such improvement in practice because it is challenging to align such conditional distributions since the target labels $\mY^{\dT}$ are unknown. More advanced approaches are often needed, which we will review  in Sec.~\ref{subsec:struc-cond-shift}. 
 Given such, keeping regularization in the last layer is often needed in practice to (approximately) achieve  $\mathbb{P}_{\phi}^{\dS} = \mathbb{P}_{\phi}^{\dT}$.


\textbf{Case II: With structure shift.} With structure shift  $\bP_{\mA\times \mY}^\dS\neq\bP_{\mA\times \mY}^\dT$, each layer of the GNN will induce distribution shift in node representations even if the distributions in the previous layer get aligned across domain, so regularization on the last-layer node representations is generally needed to achieve $\mathbb{P}_\phi^\dS = \mathbb{P}_\phi^\dT$. Then, the question in this case is that if \underline{extra regularizations} for $\mathbb{P}_{\phi_{\leq l}}^\dS = \mathbb{P}_{\phi_{\leq l}}^\dT$, for $l<L-1$ are further helpful. Unfortunately, with a simple proof, as Prop.~\ref{prop:intermediate-layer} shows, adding such regularizations will not improve objective Eq.~\eqref{eq:analysis-goal}, which thus cannot improve the bound of the error in the target domain (Eq.~\eqref{eq:bound}).  

\begin{proposition}\label{prop:intermediate-layer}
    Suppose regularization on the last-layer node representations is always adopted to achieve $\mathbb{P}_{\phi}^\dS = \mathbb{P}_{\phi}^\dT$. Then, adding regularization to the intermediate node representations
    $\mathbb{P}_{\phi_{\leq l}}^\dS = \mathbb{P}_{\phi_{\leq l}}^\dT$, for $l<L-1$ cannot further reduce the optimal error indicated by the objective of Eq.~\eqref{eq:analysis-goal}.
\end{proposition}



Combining Case I and Case II, we claim that optimizing the error bound  Eq.~\eqref{eq:bound} for the target domain by solving Eq.~\eqref{eq:analysis-goal} is necessary and typically optimal to regularize only the last-layer node representations to make their distributions invariant across domains. 

Although the above analysis justifies some rationale of previous GDA approaches, we observe its big limitation, that is we entirely ignore the second term in Eq.~\eqref{eq:bound}. As shown in Fig.~\ref{fig:HEPeg}, the ground-truth labeling functions in many real-world applications with graph-structured data may shift across domains. Ignoring such a shift yields suboptimal solutions. 
Our next section is to formalize the above issue and propose a principled algorithm to address it.

\vspace{-2mm}
\section{The Structural Re-weighting Algorithm}

In this section, we first introduce the issue of conditional structure shift (CSS). 
Then, we propose our structural re-weighting algorithm \proj to remove this shift for GDA. As a generic approach to align graph distributions for node classification tasks, \proj can also improve the vanilla training of GNNs and approaches for OOD generalization such as Mixup~\cite{wang2021mixup,verma2021graphmix}.     


\subsection{The Issue of Conditional Structure Shift}
\label{subsec:struc-cond-shift}



 
 The conditional shift has been recently investigated in the setting without graph structure~\cite{zhang2013domain, zhao2019learning,tachet2020domain}. It describes the label-conditional distribution of features shifts across domains, which corresponds to Attribute Shift $\bP_{X|Y}^{\dS}\neq \bP_{X|Y}^{\dT}$ in our context as defined in Sec.~\ref{subsec:shift-def}. This problem can be addressed in principle only  with some proper assumptions, e.g., the features in the target domain can be written as a location-scale transformation of the features in the source domain~\cite{zhang2013domain,gong2016domain}. Recent works have also adopted adversarial training to align the estimated conditional distributions based on pseudo labels $\hat{\mY}^{\dT}$ in the target domain~\cite{long2018conditional} or combined with instance-re-weight approaches~\cite{tachet2020domain} to address both of the issues of  conditional shift and label shift (i.e., $\bP_{Y}^{\dS}\neq \bP_{Y}^{\dT}$ by using our notation). 

However, none of the previous works have considered conditional structural shift (CSS) for graph-structured data: 
\begin{definition}[Conditional Structure Shift] 
$\bP_{\mA|\mY}^\dS\neq\bP_{\mA|\mY}^\dT$, where $\bP_{\mA|\mY}^\dU$ is a conditional distribution induced from $\bP_{\mA\times\mY}^\dU=\bP_{\mA|\mY}^\dU\bP_{\mY}^\dU$ . 
\end{definition}

According to the definition, the structure shift defined in Sec.~\ref{subsec:shift-def} may be caused by either CSS or label shift. Here, we study CSS, as it happens a lot in real-world graph data but cannot be addressed by simply extending previous methods. We leave its combination with label shift $\bP_{Y}^{\dS}\neq \bP_{Y}^{\dT}$ and attribute shift $\bP_{X|Y}^{\dS}\neq \bP_{X|Y}^{\dT}$ for the future studies. 

We first use an example to show the sub-optimality of previous GDA methods as their goal of pursuing domain-invariant distributions of node representations. We are inspired by the observation in Fig.~\ref{fig:HEPeg} and propose the following example with CSS based on the Contextual Stochastic Block Model (CSBM)~\cite{deshpande2018contextual}. 

\begin{definition}[Contextual Stochastic Block Model] 
    \label{def:CSBM}
    CSBM is the model that combines the stochastic block model and node attributes for the random graph generation. CSBM with nodes from $k$ classes is defined with parameters $(n, \mB, \mathbb{P}_0, \dots, \mathbb{P}_{k-1})$. Here, $n$ is the number of nodes. $\mB$ is a $k\times k$ edge connection probability matrix. $\mathbb{P}_i$, $0\leq i< k$, characterizes the distribution of node attributes of a node  from class $i$. For any node $u$ from class $i$ and any node $v$ from class $j$ in a graph generated from the model, the probability of an edge connecting them is denoted by $B_{ij}$, an entry of $\mB$. $\mB=\mB^\top$ for undirected graphs. 
    For the CSBM, all node attributes and edges are generated independently given node labels.
\end{definition}

\begin{example}
\label{eg}
Suppose graphs in the source and target domains are generated from  CSBM$(n,\mB^{\dS}, \bP_{0}, \bP_{1})$ and CSBM$(n,\mB^{\dT}, \bP_{0}, \bP_{1})$, respectively. Suppose either class in either model contains $n/2$ nodes. With some constants $p,r\in (0,1/2)$ and $\delta \in [-p, p]/\{0\}$, for $i\in\{0,1\}$, let $\bP_{i}(X) = r$ if $X=i$ and $\bP_{i}(X) = 1-r$ if $X$ is \texttt{M.V.} (denoting a default value other than 1 or 0), and 
\begin{equation}
    \mB^{\dS}=\left[\begin{array}{cc}
    p & p \\ 
    p & p-\delta \end{array}\right], 
    \mB^{\dT}=\left[\begin{array}{cc}
    p+\delta & p \\ 
    p & p \end{array}\right],
\end{equation}
So, there is no label shift or attribute shift but contains CSS. The nodes with attribute \texttt{M.V.} on the graphs generated from the above two CSBMs are used to formulate the training and test datasets, respectively. 
\end{example}
Given this example, we can quantitatively show the suboptimality of using a single shared encoder $\phi$ to learn domain-invariant node representations 
in the following proposition. 
\begin{proposition} \label{prop:example}
One-layer GNNs are adopted to solve the GDA task in Example~\ref{eg}. By imposing $\mathbb{P}_{\phi}^{\dS} = \mathbb{P}_{\phi}^{\dT}$ through a GNN encoder $\phi$ shared across the two domains, the classification error in the target domain $\epsilon_{\dT}(g,\phi) \geq 0.25$, while if without such a constraint, there exists a GNN encoder $\phi$ such that  $\epsilon_{\dT}(g,\phi)\rightarrow 0$ as $n\rightarrow \infty$.
\end{proposition}

\begin{algorithm}[t]
\caption{\proj with different training pipelines}
\label{method:alg_table}
\begin{algorithmic}[1]
\STATE \textbf{Input} One or several source graphs $\mathcal{G}_\dS$ with node labels $\mY_\dS$; One or several target graphs $\mathcal{G}_\dT$; A GNN $\phi$, a domain discriminator $q$, and a classifier $g$; The total epoch number $n$, the epoch index $m$ to start \proj, the epoch period $t$ for weight update and $\lambda$.
\WHILE {epoch $< n$ \text{or} \,\text{not converged}} 
    \IF {epoch $\geq$ $m$ } 
        \STATE When epoch $\equiv m\, (\text{mod}\,t)$, get target node  \\ representations $\mh_\dT=\phi(\dG_\dT)$, and update \\ estimation $\hat{\mB}^\dT$ with $\hat{\mY}_\dT = g(\mh_\dT)$ (Eq.~\eqref{eq:csbm-para})
        \STATE Add edge weights to $\mathcal{G}_\dS$ according to \\ $(1-\lambda)\mathbf{1}\mathbf{1}^\top + \lambda\hat{\mB}^\dT ./ \mB^\dS$
    \ENDIF
    \STATE  Get $\mh_\dS=\phi(\mathcal{G}_\dS)$, $\hat{\mY}_\dS=g(\mh_\dS)$ in the source domain
    \STATE  \textbf{Case 1: \projad}
    \begin{ALC@g}
        \STATE  Update $\phi,q$ via $\min_{q}\max_{\phi}\mathcal{L}_{\text{ADV}}(q(\mh_\dS), q(\mh_\dT))$ \\
    \end{ALC@g}
    \STATE  \textbf{Case 2: \projmix}
    \begin{ALC@g}
        \STATE  Get mixed-up predictions $\hat{\mY}_\dS$ and labels $\mY_\dS$ 
    \end{ALC@g}
    \STATE  \textbf{Case 3: \projerm} 
    \begin{ALC@g}
        \STATE Nothing to do
    \end{ALC@g}
    \STATE  Update $\phi$ and $g$ as $\min_{\phi, g}\mathcal{L}_{\text{ERM}}(\hat{\mY}_\dS,\mY_\dS)$, 
\ENDWHILE
\end{algorithmic}
\end{algorithm}


\subsection{\proj to Reduce Conditional Structure Shift}
\label{subsec:detailrw}

The previous example inspires our algorithm \proj to address CSS for node classification tasks. Note that one layer of message passing in a GNN (Eq.~\eqref{eq:GNN}) encodes the information of a tuple $(h_v^{(l)},\Xi_{\dN_v}^{(l)})$, where $\Xi_{\dN_v}^{(l)}=\ldbb h_u^{(l)}|u \in \dN_v\rdbb$ denotes the multiset of the representations of the neighbors. The graph structure here determines the cardinality of the multiset $\Xi_{\dN_v}^{(l)}$  and the distribution of the elements in $\Xi_{\dN_v}^{(l)}$. Our key idea is to down-sample or re-sample the elements in such multisets (i.e., bootstrapping) from the source domain so that the distribution of such multi-sets can (approximately) match that in the target domain.

Specifically, consider the first layer of a GNN $\phi$ that runs on graphs sampled from $k$-class CSBM$(n, \mB^{\dU}, \bP_{0},..., \bP_{k-1})$ for domain $\dU\in\{\dS,\dT\}$. Here, $\mB^{\dS}\neq \mB^{\dT}$, which indicates that there exists a CSS comparing a class-$i$ node $v$ in the target domain and a class-$i$ node $v'$ in the source domain. In the multiset $\Xi_{\dN_v}^{(0)}$ (or $\Xi_{\dN_{v'}}^{(0)}$), there will be in expectation $nB_{ij}^{\dT}$ (or $nB_{ij}^{\dS}$ resp.) many node attributes sampled from $\bP_j$ for $j\in [k]$. 
Therefore, to align the cardinality and the distribution of elements of the multiset $\Xi_{\dN_{v'}}^{(0)}$ with those of $\Xi_{\dN_v}^{(0)}$, we propose to resample (if $B_{ij}^{\dT}> B_{ij}^{\dS}$) or downsample ($B_{ij}^{\dT}<B_{ij}^{\dS}$) the elements of the class-$j$ neighbors of $v'$ to $nB_{ij}^{\dT}$ many. The following-up layers adopt the same sampling strategy. 

In practice, GNNs often adopt sum/mean pooling (also in our experiments) to aggregate these multisets. Then, the above sampling strategy reduces to adding a weight for each element in the source domain during message aggregation. The weight is $B_{ij}^{\dT}/B_{ij}^{\dS}$ for the element  passed from a class-$j$ node to a class-$i$ node. For other aggregation methods, a similar type of analysis can be adopted to determine the weights. To compute such weights, $\mB^{\dS}$ can be estimated based on Eq.~\eqref{eq:csbm-para} by using the node labels in the source domain. To estimate $\mB^{\dT}$, we propose to use the pseudo labels  estimated by the model during the training process, i.e., using $(\hat{y}_u,\hat{y}_v)$ instead of  $(y_u,y_v)$ in Eq.~\eqref{eq:csbm-para}. 
\vspace{-2mm}
\begin{align}\label{eq:csbm-para}
B_{ij} = \frac{|\{e_{uv}\in \dE |y_u = i, y_v = j\}|}{|\{v\in \dV |y_v = i\}|\times |\{v\in \dV |y_v = j\}|}.
\end{align}
\vspace{-2mm}

As the edge weights are based on the estimation of pseudo labels in practice that may have errors, we introduce a hyperparameter $\lambda$ to control the degree of reliance on this weight, i.e., the weight to be used in practice follows $(1-\lambda) + \lambda * B_{ij}^{\dT}/B_{ij}^{\dS}$.

Furthermore, to better understand model performance in practice, we would like to quantify the degree of CSS in each real dataset to help better understand the model performance. The metric we developed is as follows: 
\begin{align}\label{eq:CSS}
& \widehat{\text{CSS}} = \frac{1}{k*k}\sum_{i,j}\Delta B_{ij},\,\text{where}\\
& \Delta B_{ij} = \frac{1}{2}\left(\frac{|B_{ij}^\mathcal{S} - B_{ij}^\mathcal{T}|}{B_{ij}^\mathcal{S}} + \frac{|B_{ij}^\mathcal{S} - B_{ij}^\mathcal{T}|}{B_{ij}^\mathcal{T}}\right).
\end{align}
where $k$ is the number of classes. This metric measures the relative level of difference between the edge connection probability matrix, which reflects the degree of CSS. There is no CSS when the metric is equal to 0. We calculate the degree of CSS for each real dataset we use for experiments in Table ~\ref{table:CSS}.

Lastly, we should note that the above analysis has limitations. 
First, we did not consider attribute shift. Attribute shift, if exists, can often be (approximately) addressed by traditional DA approaches to handle conditional shift for non-graph data~\cite{long2018conditional,tachet2020domain}. 
In our experiments, we have not tried these more advanced approaches 
but our methods have already outperformed the baselines. Second, the above analysis is based on CSBM, so the derived weights are shared across the edges when the pairs of the labels of the two end nodes are the same. We believe this constraint can be further relaxed and improved. 
 \vspace{-1mm}
\subsection{\proj Combined with Different Approaches}
\label{subsec:combination}
\vspace{-0.5mm}
\proj is a generic approach to reduce CSS and should be widely applicable. Therefore, we combine \proj with three different GNN training pipelines, including \projad with adversarial-based training~\cite{ganin2016domain}, \projmix with mixup training on graphs~\cite{wang2021mixup} and \projerm with vanilla GNN training. These different combinations can be viewed as options that handle the attribute shift and CSS at different levels that vary across applications. 
For instance, \projerm or \projmix often performs well if there is no or only small attribute shift, respectively, 
while \projad will perform better with larger attribute shifts. 

The algorithm is summarized in Algorithm~\ref{method:alg_table}, where \proj is a separate module before the GNN encodes the data, which is compatible with different training pipelines. After $m$ training epochs, \proj calculates the edge weights for the source graphs to reduce CSS (lines 3-6). 
Different training pipelines may have different training losses. Besides the traditional empirical risk minimization (ERM) loss (via  $\min_{\phi,g}\mathcal{L}_{\text{ERM}}$ in Eq.~\eqref{eq:erm}) in line 14, \projad follows DANN~\cite{ganin2016domain} that trains the GNN $\phi$ and a domain discriminator $q$ (via $\max_{\phi}\min_{q}\mathcal{L}_{\text{ADV}}$ in Eq.~\eqref{eq:adv}) in line 9. Adversarial training comes into play where $q$ tries to correctly identify the source and target samples, while $\phi$ seeks to align the distributions of the  source and target samples to confuse $q$. \vspace{-0.5mm} 
\begin{align}
&\mathcal{L}_{\text{ERM}} \triangleq \sum_{u\in \dV_{\dS}} \text{cross-entropy}(y_{v},g(h_v)) \label{eq:erm} \\ 
&    \mathcal{L}_{\text{ADV}} \triangleq -(\sum_{u\in \dV_{\dS}} \log[q(h_u)] + \sum_{u\in \dV_{\dT}}\log[1-q(h_v)]) \label{eq:adv}
\end{align}
where $h_u, h_v$ are node  from $\mh_\dS$ and $\mh_\dT$. \projmix also adopts the loss   $\min_{\phi,g}\mathcal{L}_{\text{ERM}}$ while the output  $\mh_\dS$ and label $\mY$ for loss calculation are the post-mixup features and labels. The details can be found in~\cite{wang2021mixup}.

\vspace{-1mm}
\section{Experiments}
\label{sec:exp}
We evaluate \proj with the combination with the three training pipelines introduced in Sec.~\ref{subsec:combination} and compare them with existing GDA and Graph OOD baselines. The experiments are done on one synthetic dataset, one real dataset from the HEP scientific application, and four real-world benchmark networks under various types of distribution shifts. We will briefly introduce the datasets, baselines, and experiment settings. More details such as the statistics of the datasets and hyperparameter tuning can be found in Appendix~\ref{apd:exp}.


\begin{table*}[t]
\vspace{-1.8mm}
\caption{Synthetic CSBM results. The \textbf{bold} font and the \underline{underline} indicate the first and second best model respectively, $\dagger$ indicates the significant improvement, where the mean-1*std of a method $>$ the mean of its corresponding backbone model.}
\vspace{-0.5cm}
\begin{center}
\begin{adjustbox}{width=0.95\textwidth}
\begin{small}
\begin{sc}
\begin{tabular}{lcccccc}
\toprule
  &   $q=0.016$                 &    $q=0.014$                &      $q=0.012$             & $q=0.01$                  & $q=0.006$                  & $q=0.001$                  \\
\midrule
ERM & $36.52\pm3.76$ & $41.62\pm5.92$ & $48.66\pm6.31$ & $57.29\pm5.28$ & $89.72\pm2.62$ & $100\pm0$ \\
DANN  & $64.25\pm5.69$ & $72.56\pm8.54$ & $79.63\pm6.84$ & $86.29\pm8.14$ & $96.88\pm1.35$ & $100\pm0$ \\
CDAN    & $67.53\pm4.98$ & $75.38\pm7.46$ & $82.51\pm6.95$ & $89.73\pm7.44$& $97.03\pm1.09$ & $100\pm0$ \\
UDAGCN          & $51.98\pm1.31$ & $57.83\pm3.05$ & $59.74\pm1.52$ & $65.97\pm1.66$ & $98.25\pm0.52$ & $100\pm0$ \\
EERM          & $57.36\pm4.52$ & $65.88\pm3.09$ & $70.12\pm10.26$ & $72.87\pm13.70$ & $95.01\pm3.88$ & $100\pm0$ \\
Mixup          & $62.54\pm2.77$ & $69.21\pm2.03$ & $74.92\pm1.56$ & $82.87\pm3.45$ & $96.89\pm0.38$ & $100\pm0$ \\
\midrule
\projerm  & $85.24^\dagger\pm1.63$ & $87.92^\dagger\pm1.77$&$90.26^\dagger\pm1.05$ &$93.84^\dagger\pm0.98$ & ${98.28}^\dagger\pm0.14$&$\mathbf{100}\pm 0$\\
\projad    & $\underline{86.37}^\dagger\pm3.92$ & $\underline{89.22}^\dagger\pm1.83$ & $\underline{91.53}^\dagger\pm2.41$ & $\underline{94.08}^\dagger\pm0.98$ & $\mathbf{98.40}^\dagger\pm0.34$ & $\mathbf{100}\pm0$ \\
\projmix    & $\mathbf{88.48}^\dagger\pm1.93$ & $\mathbf{89.76}^\dagger\pm1.15$ & $\mathbf{92.08}^\dagger\pm1.13$ & $\mathbf{94.26}^\dagger\pm0.99$ & $\underline{98.35}^\dagger\pm0.23$ & $\mathbf{100}\pm0$ \\

\bottomrule
\label{table:CSBM}
\end{tabular}
\end{sc}
\end{small}
\end{adjustbox}
\end{center}
\vspace{-8.5mm}
\end{table*}

\vspace{-1mm}
\subsection{Datasets} 

\textbf{CSBM} is the synthetic dataset we use that consists of graphs generated from 3-class CSBMs. Each class in each graph contains 1000 nodes. We do not consider attribute shift but only structure shift to directly demonstrate the effectiveness of \proj. The node attributes in three classes in both domains satisfy Gaussains $\mathbb{P}_0 = \dN([-1, 0], I), \mathbb{P}_1 = \dN([1, 0], I), \mathbb{P}_2 = \dN([3, 2] , I)$. The intra-class edge probabilities are both 0.02 for the two domains. The inter-class edge probability ($q$ in table~\ref{table:CSBM}) in the target domain is 0.002 while that in the source domain varies from 0.001 to 0.016.

\textbf{DBLP and ACM} are two paper citation networks obtained from DBLP and ACM respectively. Each node represents a paper, and each edge indicates a citation between two papers. The goal is to predict the research topic of a paper. Here, we train the GNN on one network and test it on the other, which is denoted by $D\rightarrow A$ or $A \rightarrow D$. 
The original networks are provided by ArnetMiner~\cite{tang2008arnetminer}. We use the processed versions from~\cite{wu2020unsupervised}.   

\textbf{Arxiv} introduced in~\cite{hu2020open} is another citation network between all Computer Science (CS) Arxiv papers from 40 classes on different subject areas. 
Attributes are the embeddings of words in  titles and abstracts. 
The domain can be split based on either publication times or node degrees. 
For evaluation with different levels of publication time shift, we use papers published between 2018 to 2020 to test while using papers published in other time periods for training: \textbf{Time 1} is from 2005 to 2007 and \textbf{Time 2} is from 2011 to 2014. We follow~\cite{gui2022good} to partition the network into two domains based on node degrees. 

\textbf{Cora} is the fourth citation network with 70 classes~\cite{bojchevski2018deep}. 
Two domain splits are considered, named \textbf{Word} and \textbf{Degree}. The \textbf{Word} split is based on the diversity of words of a paper and the \textbf{Degree} split is based on node degrees, where we follow~\cite{gui2022good}.

\textbf{Pileup Mitigation} is a dataset to evaluate the approaches for a critical data processing step in HEP named pileup mitigation~\cite{bertolini2014pileup}. Particles are generated by the proton-proton collisions in the Large Hadron Collider with primary collisions (LC) and nearby bunch crossings (OC). There are multiple graphs used for training and testing. Each graph corresponds to a beam of proton-proton collisions. The particles generated from the collisions give the nodes in the graph. We connect the particles with edges if they are close in the $\eta-\phi$ space as shown in Fig.~\ref{fig:HEPeg}. As mentioned in the introduction, the task is to identify whether a neutral particle is from LC or OC. The labels of charged particles are often known. In this application, the distribution shifts may come from two sources, the shift of the types of particle decay between $pp\rightarrow Z(\nu\nu)+$  and $pp\rightarrow gg$ ~\cite{martinez2019pileup} generated from LC (mostly attribute shift with slightly structural shift), and the shift of pile-up (PU) conditions (mostly structural shift). PU$k$ means the number of collisions in the beam other than LC is $k$, where our dataset includes the cases $k\in\{10,30,50,140\}$.

\begin{table*}[t]
\vspace{-0.3mm}
\caption{Performance on real datasets. The \textbf{bold} font and \underline{underline} indicate the first and second best model respectively, $\dagger$ indicates the significant improvement, where the mean-1*std of a method $>$ the mean of its corresponding backbone model.}
\vspace{-0.5cm}
\begin{center}
\begin{adjustbox}{width=1\textwidth}
\begin{small}
\begin{sc}
\begin{tabular}{lccccccc}
\toprule
 & \multicolumn{2}{c}{DBLP and ACM} & \multicolumn{2}{c}{Cora} &   \multicolumn{3}{c}{Arxiv}\\
 Domains  & $A\rightarrow D$ & $D\rightarrow A$ &   Word &  Degree  &     time1                 & time2     & Degree       \\
\midrule
ERM & $62.48\pm3.58$ &$64.70\pm1.18$ & $64.35\pm0.44$ & $53.28\pm0.38$ & $28.08\pm0.24$  & $49.52\pm0.22$ & $57.41\pm0.14$\\
DANN  & $59.02\pm7.79$ & $65.77\pm0.46$ & $63.92\pm 0.70$ & $49.61\pm0.74$ & $24.33\pm1.19$ & $48.67\pm0.37$ & $56.13\pm0.18$ \\
CDAN    &$60.56\pm4.38$ & $64.35\pm0.83$& $62.46\pm0.94$&$52.50\pm0.96$ & $25.85\pm1.15$ & $49.22\pm0.75$ & $56.43\pm0.45$\\
UDAGCN  &$59.62\pm2.86$ &$64.74\pm2.51$ &$64.23\pm2.19$ & $58.37\pm0.72$ & $25.64\pm3.04$ &  $48.84\pm1.48$ & $55.77\pm0.83$\\
EERM         & $40.88\pm 5.10$& $51.71\pm 5.07$ & $67.43\pm 2.86$& $\underline{58.63}\pm1.12$ & OOM & OOM & OOM \\
MIXUP         & $49.93\pm0.89$ &$63.36\pm0.66$ & $\mathbf{67.73}\pm0.38$ & $58.18\pm0.52$ & $28.04\pm0.18$ & $49.98\pm0.34$ & $\underline{59.22}\pm0.22$\\
\midrule
\projerm & $\mathbf{70.19}^\dagger\pm2.10$ & $65.07\pm1.98$ & $64.34\pm 0.43$ &$55.27^\dagger\pm0.48$ & $\mathbf{28.46}^\dagger\pm0.18$  &  $48.78\pm0.40$ & $57.45^\dagger\pm0.15$\\
\projad & $\underline{66.56}^\dagger\pm 9.44$ & $\mathbf{66.57}^\dagger\pm0.42$ & $63.92\pm 0.75$ & $52.69^\dagger\pm0.36$ & $24.35\pm1.25$& $\underline{49.01}\pm0.38$& $56.36^\dagger\pm0.22$\\
\projmix    & $50.42\pm1.13$ & $\underline{66.33}^\dagger\pm0.91$ & $\mathbf{67.73}\pm0.39$ &$\mathbf{60.37}^\dagger\pm0.39$ & $\underline{28.28}^\dagger\pm0.52$  & $\mathbf{50.34}^\dagger\pm0.31$ & $\mathbf{59.99}^\dagger\pm0.09$ \\

\bottomrule
\label{table:real}
\end{tabular}
\end{sc}
\end{small}

\end{adjustbox}
\end{center}
\vspace{-7mm}
\end{table*}

\subsection{Baselines and Settings}
\textbf{Baselines} 
\proj is combined with the training pipelines of adversarial training, mixup and ERM. Therefore, we choose the corresponding baselines DANN~\cite{ganin2016domain}, graph Mixup~\cite{wang2021mixup} and the vanilla ERM with GCN~\cite{kipf2017semisupervised} as the backbone for direct comparisons. We also adopt UDAGCN~\cite{wu2020unsupervised}, EERM~\cite{wu2022handling} and CDAN~\cite{long2018conditional} with the same backbone for further comparisons. CDAN was proposed to handle the conditional shift and the label shift of the distributions of last-layer node representations. We choose GCN as most baselines use this backbone in their original literature. 



\begin{table*}[t]
\caption{HEP dataset with different PU conditions and Physical process. The \textbf{bold} font indicate the best model, $\dagger$ indicates the significant improvement, where the mean-1*std of a method $>$ the mean of its corresponding backbone model.}
\vspace{-5mm}
\begin{center}
\begin{adjustbox}{width=0.96\textwidth}
\begin{small}
\begin{sc}
\begin{tabular}{lcccccc}
\toprule
  & \multicolumn{4}{c}{PU Conditions} &  \multicolumn{2}{c}{Physical processes}\\
 Domains  &   $\text{PU}30 \rightarrow 10$    & $\text{PU}10 \rightarrow 30$ & PU$140\rightarrow 50$ & PU$50\rightarrow 140$ & $gg \rightarrow Z(\nu\nu)$      &     $Z(\nu\nu) \rightarrow gg$      \\
\midrule
ERM & $69.83\pm0.43$ & $70.73\pm0.46$ &$68.70\pm0.56$ & $68.28\pm 0.65$& $63.09\pm0.48$ & $66.53\pm1.04$ \\
DANN  & $70.14\pm0.52$ & $71.29\pm0.58$ & $69.01\pm 0.42$ & $68.98\pm 0.63$& $63.15\pm0.66$ & $66.24\pm0.97$ \\
\midrule
\projerm & $71.35^\dagger\pm0.76$& $71.95^\dagger\pm0.24$ & $69.43^\dagger\pm0.65$ & $69.05\pm0.36$ & $63.55\pm0.40$& $\mathbf{67.73}\pm0.93$\\
\projad    & $\mathbf{70.77}^\dagger\pm0.52$ & $\mathbf{71.96}\pm0.73$ &$\mathbf{69.88}^\dagger\pm 0.71$ & $\mathbf{70.54}\pm0.84$& $\mathbf{64.36}^\dagger\pm0.58$& $66.91\pm0.67$ \\

\bottomrule
\label{table:Pileup}
\end{tabular}
\end{sc}
\end{small}
\end{adjustbox}
\end{center}
\vspace{-9mm}
\end{table*}

\textbf{Settings and Metric.}
By the definition of GDA, the graphs in the source domain are used for training, while the graphs in the target domain are used for validation and testing. Specifically, we use 20 percent of node labels in the target domain for validation, and the rest 80 percent are held out for testing. 
The estimation of $\hat{\mathbf{B}}^{\mathcal{T}}$ in the target domain for \proj uses the ground-truth labels of the target validation nodes (as assumed to be known) and the pseudo labels for the hold-out target testing nodes. The final evaluation scores included in the tables are based on the accuracy score for the node classification tasks on the hold-out target testing nodes. The selection of the best model is based on the score on the target validation nodes. All results are summarized based on 5 times independent experiments.


\subsection{Result Analysis}

The experiment results over the synthetic datasets are in Table~\ref{table:CSBM}. 
As the performance of ERM shows, CSS may cause significant performance decay. 
All baseline methods can deal with CSS to some extent while still performing significantly worse than \proj-based approaches. 
Also, the improvement of \proj increases with how much CSS the data holds. Particularly, \proj is able to boost the performance by more than 20\% over the best baseline. The results match our expectations well since the synthetic datasets are precisely aligned with the motivation of \proj. 

Table~\ref{table:real} includes the results for four real-world citation datasets. For all the datasets, \projerm, \projad, and \projmix outperform their corresponding baseline models ERM, DANN, and Mixup, respectively. Moreover, across all the datasets, one of \projerm, \projad and \projmix achieves the best performance, and over six of the seven settings, \proj based methods have achieved significant improvement, i.e., the differences in means greater than one times the std of our models. Note that it is hard to expect a significant improvement of \proj in the GDA setting without much CSS, e.g., the setting of \textbf{Word} (\textbf{Cora}) whose distribution shift is mostly due to attribute shift. In comparison, in the settings of \textbf{Degree} (\textbf{Cora}) and \textbf{Degree} (\textbf{Arxiv}), and \textbf{DBLP and ACM}, the improvements based on reweighting are more significant. The results match our intuition and are supported by the quantitative CSS we calculated in Table \ref{table:CSS}. Over the datasets with larger CSS scores, 
\proj demonstrates more significant improvement over the baselines. The \proj-based methods performance largely relies on the corresponding baseline performances.  \projad tends to be less stable and works better when there is a large distribution shift. \projerm and \projmix are much more stable and have close performances when the distribution shift is small. 

Finally, for the \textbf{HEP} datasets, we compare \projerm and \projad with the corresponding baselines ERM and DANN. Note that the current pipelines of \projmix and Mixup are not suitable for this dataset as these HEP datasets contain multiple graphs for either training or testing since how to properly mix up node attributes across graphs needs a non-trivial design, which is left for future study. A similar issue comes with other baselines such as UDAGCN originally proposed for single graphs used for training and testing.
Under the domain shift caused by different PU conditions, we have often observed significant improvements over the case adapting from the higher PU levels to lower PU levels, while when being trained on lower PU levels and tested on higher PU levels, there are some but marginal improvements. These results match previous findings in the studies on this HEP application with ML technique\cite{li2021semisupervised, komiske2017pileup}. We suspect the reason is that the model learned with low PU levels tends to be more robust to the distribution shift. \proj-based methods also help with the cases with shifts in particle types, although the improvements are not significant. 

Besides the difficulty of the physics task itself that causes marginal performance in absolute accuracy scores, we suspect two additional reasons that may diminish the \proj performance for HEP datasets. The first reason is that this pileup mitigation task is a binary classification, which is often easier than multi-class classification tasks due to the simpler decision boundary. The second reason may come from the multi-graph training and testing procedure, where the average overweight calculations in \proj technique can limit the model performance. 

\begin{figure}[t]
    \centering
    \includegraphics[width=0.49\linewidth]{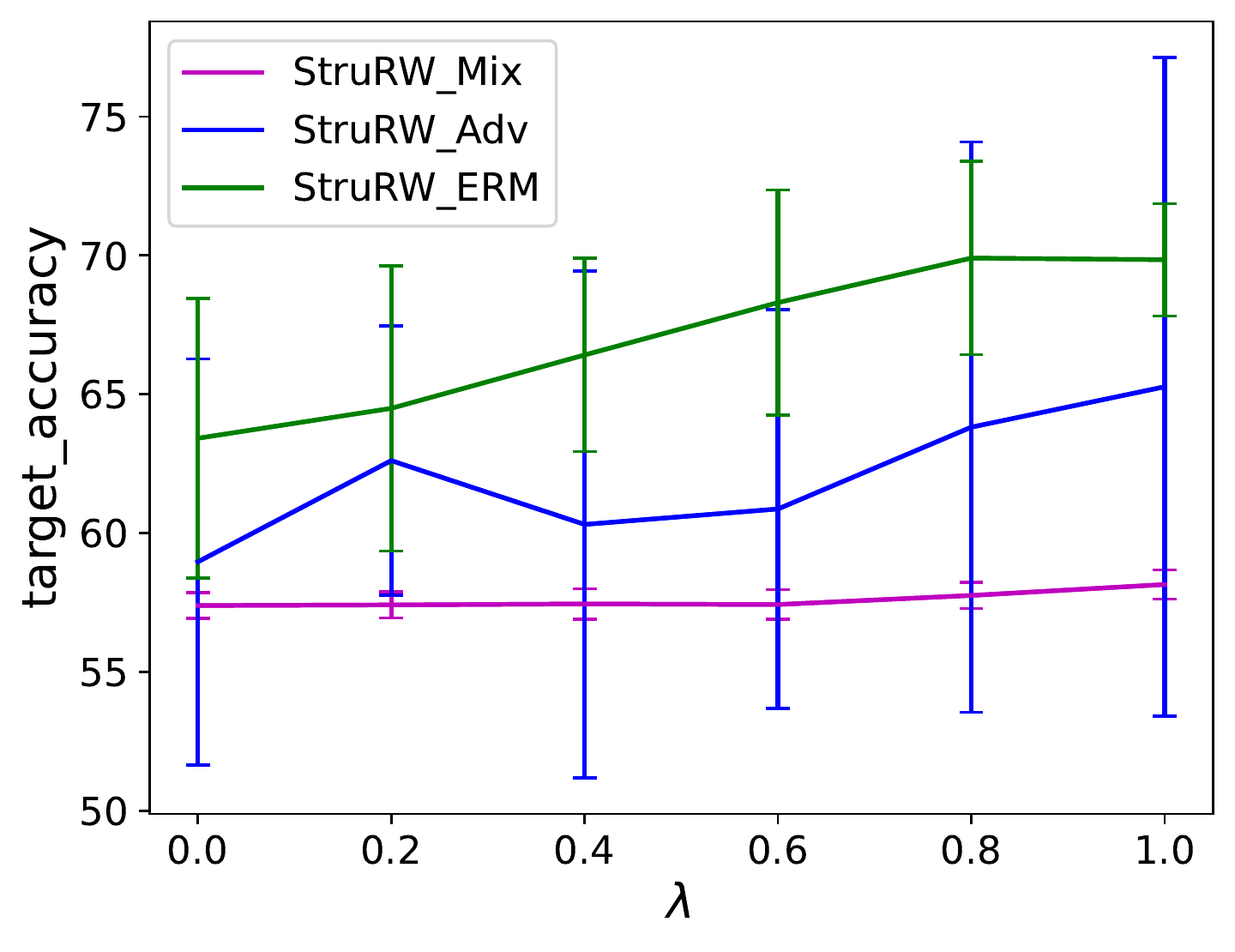}
    \includegraphics[width=0.49\linewidth]{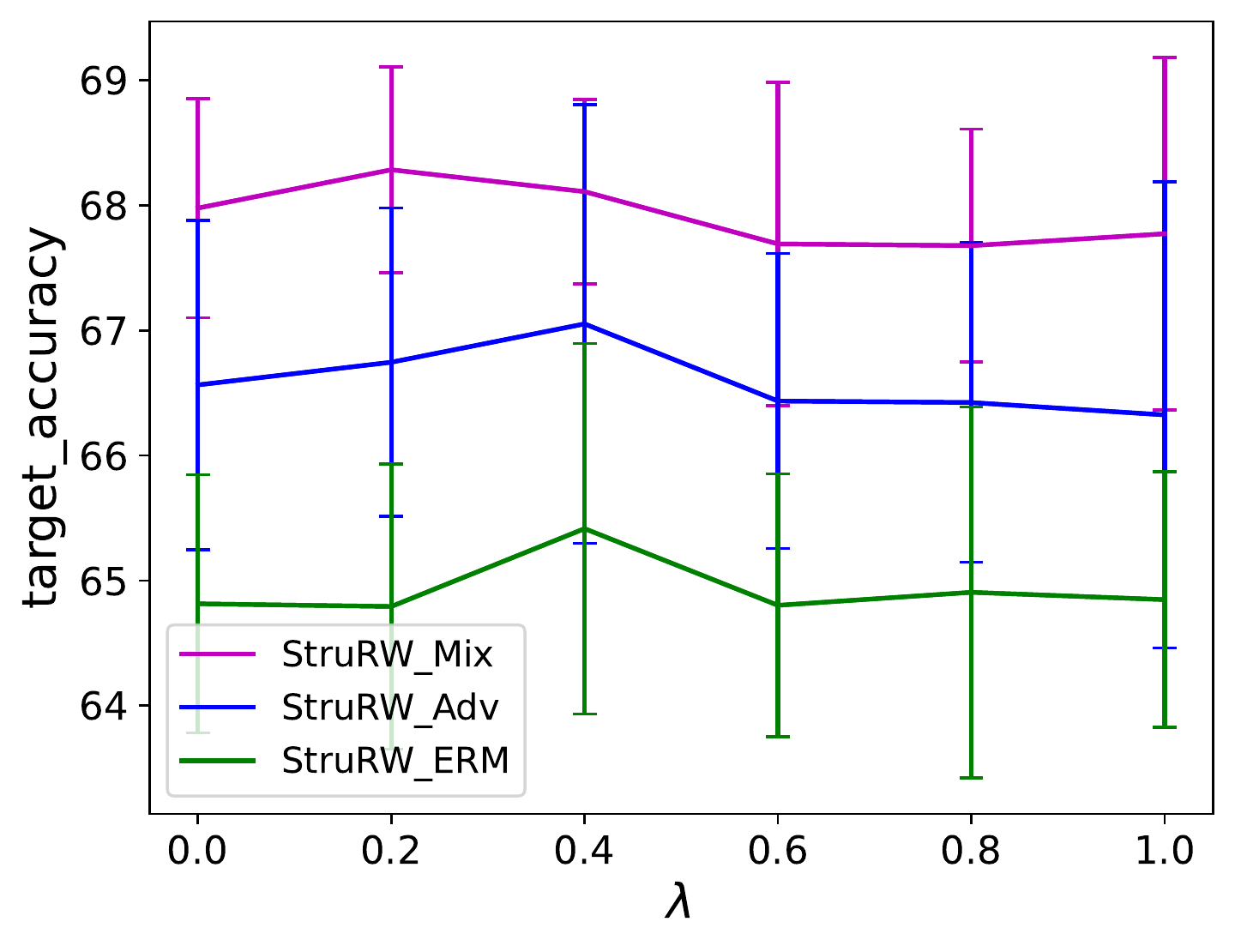}
\vspace{-3mm}
\caption{The hyperparameter study of $\lambda$ for three StruRW models over the datasets ACM$\rightarrow$DBLP and DBLP$\rightarrow$ACM.}
\label{fig:lambda}
\vspace{-6mm}
\end{figure}

\textbf{Hyperparameters.} Besides the normal hyperparameter tuning including learning rate, model architecture, and epoch as some basic setups, our \proj relies on three hyperparameters: the epoch $m$ to start \proj, the time period $t$ to calculate weights, and the $\lambda$ for the degree to adopt the reweighted message. A general rule to select $\lambda$ is that if the original CSS is large, we may want to pay attention to the reweighted message more so as to alleviate the CSS. The hyperparameter study over $\lambda$ is demonstrated in Fig.~\ref{fig:lambda} under the settings with ACM $\rightarrow$ DBLP and DBLP $\rightarrow$ ACM. 
The specific values of these hyperparameters and some baselines hyperparameters are reported in Appendix~\ref{apd:exp}.

\section{Conclusion}
This work studies graph domain adaptation for node classification problems. We analyze the effects of different types of distribution shifts in graph-structured data. We have shown the advantages of the common solution to align last-layer node representations for GDA while disclosing the issues of using a shared GNN encoding pipeline to achieve so. We show that such a limitation can be caused by a newly identified type of distribution shift, named conditional structural shift, which widely shows up in practice. To reduce CSS in the data, we have proposed a new approach \proj that asks to reweight the graphs in the source domain during GNN encoding. Extensive evaluation over synthetic graphs, real-world graphs, and the pileup-mitigation application in HEP has demonstrated the effectiveness of \proj.      

\section*{Acknowledgement}
We greatly thank all the reviewers for their valuable feedback and thank Mia Liu for discussing relevant applications. S. Liu, T. Li, and P. Li are partially supported by NSF award OAC-2117997. Q.Qiu is partially supported by NIH. The work of HZ was supported in part by the Defense Advanced Research Projects Agency (DARPA) under Cooperative Agreement Number: HR00112320012, a Facebook Research Award, and Amazon AWS Cloud Credit. YF and NT are supported by Fermi Research Alliance, LLC under Contract No. DE-AC02-07CH11359 with the Department of Energy (DOE), Office of Science, Office of High Energy Physics and the DOE Early Career Research Program under Award No. DE-0000247070.

\newpage
\nocite{langley00}

\bibliography{reference}
\bibliographystyle{icml2023}

\newpage
\appendix
\onecolumn

\section{Derivation of The Error Bound in The Target Domain Eq.~\eqref{eq:bound}} \label{adp:bound}
We follow the derivation in \cite{wu2019domain}. Let $f_{\dU}^\phi(x)= \int_{x:\phi(x) = h}f_{\dU}(x)d\bP_X^\dU(x)$.
 First, we have $r_\dU(h, \phi, g) = \int_{x:\phi(x) = h}|g(h) - f_{\dU}(x)|d\bP_X^\dU(x) =|g(h) - f_{\dU}^\phi(h)|$ for $\dU\in\{\dS,\dT\}$. And thus, 

 \begin{align}\label{eq:apd-check}
 r_\dS(h, \phi, g) - r_\dT(h, \phi, g) = |g(h) - f_{\dS}^\phi(h)| - |g(h) - f_{\dT}^\phi(h)| \leq |f_{\dS}^\phi(h) - f_{\dT}^\phi(h)|.
\end{align}
Therefore, the target error can be bounded as
\begin{align}
    \epsilon_\dT(g, \phi) &= \epsilon_\dT(g, \phi) + \epsilon_\dS(g, \phi) - \epsilon_\dS(g, \phi)\\
    &=\epsilon_\dS(g, \phi) + \int_{x}|g(\phi(x))-f_\dT(x)|d\mathbb{P}_X^\dT(x) -  \int_{x}|g(\phi(x))-f_\dS(x)|d\mathbb{P}_X^\dS(x)\\
    &=\epsilon_\dS(g, \phi)+\int_h r_\dT(h, \phi, g)d\mathbb{P}_\phi^\dT(h) - \int_h r_\dS(h, \phi, g)d\mathbb{P}_\phi^\dS(h)\\
    &=\epsilon_\dS(g, \phi) + \int_h d\mathbb{P}_\phi^\dT(h)(r_\dT(h, \phi, g) - r_\dS(h, \phi, g)) + \int_h (d\mathbb{P}_\phi^\dT(h) - d\mathbb{P}_\phi^\dS(h))r_\dS(h, \phi, g)\\
      &\leq \epsilon_\dS(g, \phi) + \int_h d\mathbb{P}_\phi^\dT(h)|r_\dT(h, \phi, g) - r_\dS(h, \phi, g)| + \int_h |d\mathbb{P}_\phi^\dT(h) - d\mathbb{P}_\phi^\dS(h)|r_\dS(h, \phi, g) \\
      &\stackrel{a)}{\leq} \epsilon_\dS(g, \phi) + \int_h  d\mathbb{P}_\phi^\dT(h)|f_{\dS}^\phi(h) - f_{\dT}^\phi(h)| + \int_h |d\mathbb{P}_\phi^\dT(h) - d\mathbb{P}_\phi^\dS(h)|r_\dS(h, \phi, g)
\end{align}
where a) uses Eq.~\eqref{eq:apd-check}. 




\section{Proof for Proposition  \ref{prop:remove-attribute-shift}}
\begin{proof}
 The initial node attributes $x_v, v\in \dV$ are independently sampled from the conditional distribution $\mathbb{P}_{X|Y}$ given the node labels $y_v$. No matter whether attribute shift $\mathbb{P}_{X|Y}^\dS \neq \mathbb{P}_{X|Y}^\dT$ exists, our condition is that the transformation $\phi_{\leq 0}$ maps the attributes $x_v$ to $h_v^{(0)} $ and satisfies $\mathbb{P}_{\phi_{\leq 0}|Y}^\dS = \mathbb{P}_{\phi_{\leq 0}|Y}^\dT$. The goal is to prove that if $\mathbb{P}_{\phi_{\leq 0}|Y}^\dS = \mathbb{P}_{\phi_{\leq 0}|Y}^\dT$, then the after the GNN, node representations will reach $\mathbb
    {P}_\phi^\dS = \mathbb{P}_\phi^\dT$. 

Since the message-passing process at each GNN layer relies on the same adjacency matrix A for neighborhood aggregation, we can prove this by induction. However, it is hard to prove $\mathbb{P}_{\phi_{\leq l}|Y}^\dS = \mathbb{P}_{\phi_{\leq l}|Y}^\dT \Rightarrow \mathbb{P}_{\phi_{\leq l+1}|Y}^\dS = \mathbb{P}_{\phi_{\leq l+1}|Y}^\dT$ because $h_v^{(l)}$'s are not independent. So, we are to consider the joint distribution of $\{h_v^{(l)}|v\in \dV\}$ and graph structure $\mA$ given node labels, i.e.,  $\mathbb{P}_{\mH^{(l)}\times\mA|Y}^\dS$, and prove $\mathbb{P}_{\mH^{(l)}\times\mA|Y}^\dS= \mathbb{P}_{\mH^{(l)}\times\mA|Y}^\dT\Rightarrow \mathbb{P}_{\mH^{(l+1)}\times\mA|Y}^\dS= \mathbb{P}_{\mH^{(l+1)}\times \mA|Y}^\dT$. If this is true, we have  $\mathbb{P}_{\mH^{(L)}\times \mA|Y}^\dS= \mathbb{P}_{\mH^{(L)}\times \mA|Y}^\dT$. By integrating over $\mathbb{P}_{\mA|Y}^\dU$, we achieve $\mathbb{P}_{\mH^{(L)}|Y}^\dS= \mathbb{P}_{\mH^{(L)}|Y}^\dT.$

First, when $l=0$, since $\mathbb{P}_{\phi_{\leq 0}|Y}^\dS = \mathbb{P}_{\phi_{\leq 0}|Y}^\dT$ and all $h_v^{(0)}$'s are mutually independent. We have $\mathbb{P}_{\mH^{(0)}|Y}^\dS = \mathbb{P}_{\mH^{(0)}|\mY}^\dT$. Also, since there is no structure shift $\mathbb{P}_{\mA|Y}^\dS = \mathbb{P}_{\mA|Y}^\dT$ and $\mA$ and $\mX$ are independent given $\mY$, we have  $\mathbb{P}_{\mH^{(0)}\times \mA|Y}^\dS = \mathbb{P}_{\mH^{(0)}\times \mA|\mY}^\dT$

For $l>0$, consider the $l$th layer of GNN that takes $h_v^{(l)}, v\in \dV$ and $\mA$ as input and follows Eq.~\eqref{eq:GNN} as:
    \begin{equation}
        h_v^{(l+1)} = \text{UDT}(h_v^{(l)}, \text{AGG}(\{\{h_v^{(l)}:u\in \dN_v\}\})).
    \end{equation}
which depends on $\mH^{(l)}$ and $\mA$. So, we have 
$\mathbb{P}_{\mH^{(l+1)}\times \mA|Y}^\dS = \mathbb{P}_{\mH^{(l+1)}|Y,\mA}^\dS  \mathbb{P}_{\mA|Y}^\dS \stackrel{a)}{=} \mathbb{P}_{\mH^{(l+1)}|Y,\mA}^\dT  \mathbb{P}_{\mA|Y}^\dS = \mathbb{P}_{\mH^{(l+1)}\times \mA|Y}^\dT$.
where a) is due to the induction condition $\mathbb{P}_{\mH^{(l)}\times\mA|Y}^\dS= \mathbb{P}_{\mH^{(l)}\times\mA|Y}^\dT$, which concludes the proof.

\end{proof}


\section{Proof for Proposition \ref{prop:intermediate-layer}}
\begin{proof}
Actually, this proposition is easy to obtain from the perspective of optimization. Since the goal is always with the constraint $\mathbb{P}_\phi^\dS = \mathbb{P}_\phi^\dT$, adding an intermediate-layer regularization, say $\mathbb{P}_{\phi_{\leq l}}^\dS = \mathbb{P}_{\phi_{\leq l}}^\dT$, which makes the optimization problem~\eqref{eq:bound} as
   \begin{equation}
        \label{eq:inter-gnn}
       \min_{\phi_{>l}, \phi_{\leq l}}\epsilon_\mathcal{S}(\phi) \quad \text{s.t.}\; \mathbb{P}_{\phi_{\leq l}}^\dS = \mathbb{P}_{\phi_{\leq l}}^\dT, \mathbb{P}_\phi^\dS = \mathbb{P}_\phi^\dT
    \end{equation} 

    Comparing the objective function and constraints from Eq.~\eqref{eq:analysis-goal} and Eq.~\eqref{eq:inter-gnn}, we find the same objective but with additional invariant representation constraints in the intermediate layer of GNN. As for both the constraints on the final layer of representations are imposed, additional constraints will only restrict the feasible region for GNN parameters to further reduce the source error. Therefore, Eq.~\eqref{eq:analysis-goal} has an optimal solution no worse than Eq.~\eqref{eq:inter-gnn} in terms of a lower source classification error, which ultimately determines the bound in Eq.~\eqref{eq:analysis-goal}.
\end{proof}

\section{Proof for Proposition~\ref{prop:example}}
Recall that node attributes in both domains follow: 
\begin{align}
    \mathbb{P}_{0}(X) = \begin{cases} r &\text{if}  \; X = 0\\
1-r & \text{if} \; X = \text{Missing Value (M.V.)}
\end{cases}, \quad  
\mathbb{P}_1(X)  = \begin{cases} r &\text{if}  \; X = 1\\
1-r &\text{if}  \; X =\text{Missing Value (M.V.)}
\end{cases}
\end{align}

To classify a node $v$ with M.V. as its attribute, if we use one-layer GNN, the classification essentially reduces to classify the multi-set $\Xi_v$ of attributes from its neighbors. Let us analyze this multi-set. 

This multi-set $\Xi_v$ has the following equivalent representation: It contains at most $n-1$ elements that have values chosen from $\{0, 1, \text{M.V.}\}$. 
Therefore, $\Xi_v$ can be represented as a 3-dim vector $(c_0,c_1,c_2)$ 
where $c_0, c_1, c_2$ represent the multiplicity of each type of element $0,1,\text{M.V.}$ in the multiset and satisfy $c_1+c_2+c_3 \leq n-1$. Our analysis is based on analyzing $\bP^{\dU}(\Xi_v = (c_0,c_1,c_2)|Y_v)$ for $\dU\in\{\dS,\dT\}$ and $Y_v\in\{0,1\}$. 

\textbf{Case 1:} Let us first prove that when we use a shared GNN encoder $\phi$ to impose $\mathbb{P}^{\dS}_{\phi} = \mathbb{P}^{\dT}_{\phi}$, $\epsilon_{\dT}(g,\phi)\geq 0.25$. 

Given a GNN model $\phi$ and the classifier $g$, we partition the feature space into the 0-space $\boldsymbol{\Xi}_0(g,\phi) = \{(c_1,c_2,c_3): g\circ\phi(c_1,c_2,c_3) = 0, c_1+c_2+c_3 \leq n-1, c_i\in\mathbb{Z}_{\geq 0}\}$ and the 1-space $\boldsymbol{\Xi}_1(g,\phi) = \{(c_1,c_2,c_3): g\circ\phi(c_1,c_2,c_3) = 1, c_1+c_2+c_3 \leq n-1, c_i\in\mathbb{Z}_{\geq 0}\}$.

Recall the CSBM models for source and target domains have structures:
\begin{equation}
    \mB^{\dS}=\left[\begin{array}{cc}
    p & p \\ 
    p & p-\delta \end{array}\right], 
    \mB^{\dT}=\left[\begin{array}{cc}
    p+\delta & p \\ 
    p & p \end{array}\right],
\end{equation}

We know $\bP^{\dS}(\Xi_v|Y_v=0) = \bP^{\dT}(\Xi_v|Y_v=1)$ because in the source domain, if for $v$ with $Y_v=0$, the edge probability between $v$ and any node with label $0$ is $p$, and the edge probability between $v$ and any node with label $1$ is also $p$. In the target domain, if for $v$ with $Y_v=1$, the edge probability between $v$ and any node with label $0$ is $p$, and the edge probability between $v$ and any node with label $1$ is also $p$. Therefore, no matter what $\phi,g$ are chosen,  $\bP^{\dS}[\boldsymbol{\Xi}_i(g,\phi)|Y=0] = \bP^{\dT}[\boldsymbol{\Xi}_i(g,\phi)|Y=1]$. Therefore, 
\[1 = \bP^{\dS}[\boldsymbol{\Xi}_0(g,\phi)|Y=0] + \bP^{\dS}[\boldsymbol{\Xi}_1(g,\phi)|Y=0]  = \bP^{\dT}[\boldsymbol{\Xi}_0(g,\phi)|Y=1] + \bP^{\dS}[\boldsymbol{\Xi}_1(g,\phi)|Y=0]\leq 2(\epsilon_{\dT}(g,\phi) + \epsilon_{\dS}(g,\phi)).\]
The last inequality is because 
\[\epsilon_{\dU}(g,\phi) = \bP^{\dU}[\boldsymbol{\Xi}_0(g,\phi)|Y=1] \bP^{\dU}[Y=1] + \bP^{\dU}[\boldsymbol{\Xi}_1(g,\phi)|Y=0]\bP^{\dU}[Y=0]\]
\[\geq \frac{1}{2}\max\{\bP^{\dU}[\boldsymbol{\Xi}_0(g,\phi)|Y=1], \bP^{\dU}[\boldsymbol{\Xi}_1(g,\phi)|Y=0]\}.\] 

So, $\epsilon_{\dT}(g,\phi) + \epsilon_{\dS}(g,\phi)\geq 0.5$. It is also a reasonable assumption that $\epsilon_{\dT}(g,\phi) \geq \epsilon_{\dS}(g,\phi)$ in practice. So, we have $\epsilon_{\dT}(g,\phi)\geq 0.25$.

\textbf{Case 2:} Case 1 implies that we should not impose domain invariant distributions via the GNN encoding process shared across domains. We may prove that if the GNN encoding process $\phi$ for the target domain can be chosen differently from that for the source domain, then there is a $\phi$ $\epsilon_{\dT}(g,\phi) \rightarrow 0$ as $n\rightarrow \infty$. Here, we assume $n$ is large enough and ignore the difference between $n$ and $n-1$.

Given a node $v$ with the multiset feature $\Xi_v=(c_0,c_1,c_2)$, suppose the GNN encoder $\phi$ follows $\phi(\Xi_v) = (c_0 - c_1)/n$. 

Recall that we have the following two cases
\begin{itemize}
    \item If $v$ is from class 0 in the target domain, $c_1\sim \text{Bin}(n/2,pr)$, $c_0\sim \text{Bin}(n/2,(p+\delta)r)$
    \item If $v$ is from class 1 in the target domain, $c_1\sim \text{Bin}(n/2,pr)$, $c_0\sim \text{Bin}(n/2,pr)$.
\end{itemize}

As $c_1$ and $c_0$ are always independent, if $v$ is from class 0 in the target domain,  $\phi(\Xi_v) = \frac{1}{n}(\sum_{i=1}^{n/2} Z_i - \sum_{i=1}^{n/2} Z_i')$, where $Z_i \sim \text{Bern}((p+\delta)r)$ and $Z_i' \sim \text{Bern}(pr)$, and all $Z_i$'s and $Z_i'$'s are independent. Here,  $\text{Bern}(\cdot)$ is the Bernoulli distribution. Therefore, using Hoeffding's inequality, we have 
\begin{align}
    \mathbb{P}\left(\phi(\Xi_v) - \mathbb{E}[\phi(\Xi_v)]< t\right) \leq \exp(-\frac{nt^2}{2})
\end{align}
If pick $t=\frac{\delta r}{4}$, $ \mathbb{P}\left(\phi(\Xi_v) < pr+\frac{\delta r}{4}\right) \leq \exp(-\frac{n\delta^2r^2}{32})$. Similarly, if $v$ is from class 0 in the target domain, we have $ \mathbb{P}\left(\phi(\Xi_v) > pr + \frac{\delta r}{4}\right) \leq \exp(-\frac{n\delta^2r^2}{32})$. 

Therefore, by setting the classifier as $g(h) = 0$ if $h > pr + \frac{\delta r}{4}$ or $1$ if $h < pr + \frac{\delta r}{4}$. Then, the error rate in the target domain will be less than $2\exp(-\frac{n\delta^2r^2}{32})$, which goes to $0$ as $n$ goes to $\infty$.

\section{Supplement for Experiments}
\label{apd:exp}


\subsection{Datasets} 
\subsubsection{Dataset Statistics for ACM, DBLP, Cora, Arxiv}
Below is the summary of our real datasets with the number of nodes, number of edges, node feature dimension and number of class labels.
\begin{table}[h!]
\caption{real dataset statistics}
\vspace{-4mm}
\begin{center}
\begin{sc}
\begin{tabular}{lcccc}
\toprule
            & ACM          & DBLP           & Cora          &Arxiv  \\
\midrule
$\#$nodes          & $7410$  & $5578   $ & $19793$     &  $169343$     \\
$\#$edges        & $22270    $ & $14682    $ & $126842$    &$2315598$        \\
Node feature dimension    & $7537    $ & $7537  $ & $8710$   &    $128$     \\
$\#$labels        & $6    $ & $6  $ & $70$    &$40$        \\
\bottomrule
\label{table:datastats}
\end{tabular}
\end{sc}
\end{center}
\vskip -0.7cm
\end{table}

\subsubsection{Details for HEP datasets} 
Next, we detail some statistics and setup for the HEP datasets
For our studies, simulated datasets have been generated of different physical processes under different pileup conditions. In this study, we
select four pileup conditions where the numbers of other interactions (nPU) are 10, 30, 50, 140 respectively, and two hard scattering signal
processes, $pp\rightarrow Z_{\nu\nu}+$ jets and $pp\rightarrow gg$ jets. Later on, we will shorten as $Z(\nu\nu)$ and $gg$ for the two signals.

These HEP datasets for pileup mitigation tasks are node classification tasks but with multiple graphs. Each node represents a particle and we construct the graph based on a threshold of the relative distance between two particles in the $\eta$ and $\phi$ space as demonstrated in fig.~\ref{fig:HEPeg}. The number of graphs we used for training is 70 and the rest of 30 are left for testing. The number of labels is 2 for all the datasets and the node feature dimension is 28. Besides, the particles can be split into charged and neutral where neutral particles do not encode ground truth label information. Under our setting, we choose to encode the ground truth of charged particles into node features so as to help with classification. The node features then contain the $\eta$, pt, pdgID one hot encoding (feature to indicate the type of particle, like Hadron and Photon), and charged label encoding. The table below includes some detailed statistics associated with this HEP dataset, which is averaged over a total of 100 graphs.
\begin{table}[h!]
\caption{HEP dataset statistics}
\vspace{-4mm}
\begin{center}
\begin{sc}
\begin{tabular}{lcccccc}
\toprule
            & $PU10\_gg$          & $PU30\_gg$           & $PU50\_gg$          &$PU50\_Z(\nu\nu)$  &  $PU140\_gg$ & $PU140\_Z(\nu\nu)$\\
\midrule
$\#$nodes    & $185.17$  & $417.84   $ & $619$     &  $570.90$ & $1569.04$ & $1602.14$     \\
$\#$edges    & $1085.17    $ & $3518.43    $ & $7169.51$    &$5894.8$    &$42321.71$ & $44070.80$    \\
LC/OC ratio  & $2.8600$ & $0.2796$ & $0.1650$ & $0.0927$ & $0.0575$ &$0.0347$\\
\bottomrule
\label{table:datastats}
\end{tabular}
\end{sc}
\end{center}
\vskip -0.7cm
\end{table}

\subsection{Hyperparameter Analysis}
In this section, we will introduce our hyperparameter analysis. As mentioned in the experiment section, our \proj mainly depends on three hyperparameters to calculate and apply the edge reweighting on the source graphs. The epoch $m$ we plan to start calculating the reweighting, the frequency we update the edge weights from the last calculation $t$, and the degree we integrate the reweighted message $\lambda$ with the original message. Based on our hyperparameter tuning process, we found $\lambda$ and starting epoch $m$ tend to be important factors that impact our reweighting performance. We may want to start the reweighting early and with low lambda to rely more on the reweighted information when the CSS shift is large and has more room for improvements. Regarding the case with small CSS, we set larger $\lambda$ and update with low frequency. 

Other important hyperparameters are associated with the coefficient $\alpha$ for the gradient reversal layer in the adversarial training pipeline \projad. The two hyperparameters we can tune are the scale added in front of $\alpha$ and the max value that $\alpha$ can take when propagating the reverse gradients. It generally helps with the stability of adversarial training. 

\textbf{Model Architecture} Our backbone model is based on GCN~\cite{kipf2017semisupervised} and for all the baselines. For the DBLP and ACM datasets, we follow the hidden dimension used in the original~\cite{wu2020unsupervised} paper two layers of GNN with hidden dimension 128, encode the embeddings into 16 and followed by the classifier with hidden dimension 40. Both the Arxiv and Cora datasets use 300 hidden dimensions with 2 GNN layers. The HEP datasets use hidden dimension 50 and CSBM adopts hidden dimension 20. 

\textbf{learning rate and epochs} We select some space to tune the learning rate, where the models mostly take the learning rate as 0.007, 0.004, and 0.001. The adversarial-based model will prefer a learning rate of 0.007 and mixup-based models will prefer a learning rate of 0.001 and 0.004. For the adversarial-based training model DANN and \projad, we set the epochs to be 300 and for the mixup model, we will take the epochs to be 200. 

\textbf{GRL coefficient $\alpha$} This value will scale the gradient when we propagate the gradient back. The original calculation is based on the epochs where $\alpha$ is equal to the current epoch divided by the total epochs. Also, it can follow the calculation implemented in DANN~\cite{ganin2016domain}. Here we add two additional hyperparameters to tune this $\alpha$ for more stable performance. One is a constant that is multiplied in front of this alpha The search space we set for this parameter is mainly $\{1, 1.5, 2\}$. The other is the max value this $\alpha$ can take, with search space $\{0.1, 0.5, 1\}$. 

\textbf{starting epoch $m$ and the \proj time period $t$} The starting epoch means that we will start imposing edge weights on the source graph after epoch $m$ and the \proj freq means we update the edge weights calculated every $t$ epochs. However, note that in the middle of the $t$ epochs, we will still keep the edge weights calculated from the last time until a new update. The search space for $m$ is $\{100, 150, 200, 250\}$ for experiments with 300 epochs and $\{50, 100, 150\}$ for epoch 200 trainings. The search space for $t$ is $\{1, 5, 10, 15\}$, we found this parameter does not affect the performance as much as the starting epoch. For the experiment that already has good ERM results with smaller shift like Cora, we tend to start later. For the cases where the effect of \proj is significant, it generally starts early at epoch 50.

\textbf{$\lambda$ in \proj} This is a ratio to guide message aggregation, 0 stands for the case that completely adopts the reweighted message and 1 corresponds to the GNN original message. It is discussed in the paper's main text and in Fig.\ref{fig:lambda} that we choose large $\lambda$ when CSS is large and small $\lambda$ when CSS is small. The specific $\lambda$ for each different dataset is shown in Table \ref{table:spec_lambda}.

\begin{table*}[t]
\caption{Optimal $\lambda$ value for each real dataset}
\vspace{-5mm}
\begin{center}
\begin{adjustbox}{width=0.96\textwidth}
\begin{small}
\begin{sc}
\begin{tabular}{lccccccc}
\toprule
 Domain Splits  &   $A\rightarrow D$    & $D \rightarrow A$ & Cora Word & Cora Degree & Arxiv time1      &     Arxiv time2 & Arxiv Degree      \\
\midrule
\projerm & $0.8$ & $1$ & $0.8$ & $0.8$& $0.1$ & $0.1$ & $0.2$\\
\projad & $1$ & $0.6$ & $0.1$ & $1$& $0.1$ & $0.1$ & $0.2$\\
\projmix & $1$ & $0.6$ & $0.6$ & $1$& $0.1$ & $0.1$ & $0.2$\\

\bottomrule
\label{table:spec_lambda}
\end{tabular}
\end{sc}
\end{small}
\end{adjustbox}
\end{center}
\vspace{-7mm}
\end{table*}

\textbf{Baseline hyperparameters} For the baseline models, we use the same GNN backbone and the same model architecture as discussed above. The baseline DANN, ERM and Mixup share the same set of hyperparameters as \projad, \projerm and \projmix respectively. For the UDAGCN baseline, we keep the original set of hyperparameters published in their work. For EERM baselines, I kept the original setting suggested in their paper.

\end{document}